\pdfoutput=1

\documentclass[11pt]{article}

\usepackage[]{acl}

\usepackage{times}
\usepackage{latexsym}

\usepackage[T1]{fontenc}

\usepackage[utf8]{inputenc}

\usepackage{microtype}

\usepackage{amsmath,amsfonts,bm}

\def\eqref#1{equation~\ref{#1}}

\def\1{\bm{1}}

\DeclareMathAlphabet{\mathsfit}{\encodingdefault}{\sfdefault}{m}{sl}
\SetMathAlphabet{\mathsfit}{bold}{\encodingdefault}{\sfdefault}{bx}{n}

\usepackage{hyperref}
\usepackage{url}

\usepackage{utfsym}
\usepackage{booktabs}
\usepackage{amsfonts}
\usepackage{nicefrac}
\usepackage{microtype}
\usepackage{graphicx}
\usepackage{enumitem}
\usepackage{subcaption}
\usepackage[export]{adjustbox}
\usepackage{amsmath}
\usepackage{tikz}
\usepackage[most]{tcolorbox}
\usepackage{bbm}
\usepackage{caption}
\usepackage{array}
\usepackage{soul}
\usepackage[super]{nth}
\usepackage{float}
\usepackage{wrapfig}
\usepackage{mdframed}
\usepackage{multirow}
\usepackage{fontawesome}
\usepackage{mathtools}
\usepackage{amstext}
\usepackage{multicol}

\newcommand{\orange}[1]{\textcolor{orange}{#1}}
\newcommand{\red}[1]{\textcolor{red}{#1}}
\newcommand{\blue}[1]{\textcolor{blue}{#1}}

\title{Reasons to Reject? \\ Aligning Language Models with Judgments}

\author{
  Weiwen Xu$^{\heartsuit\spadesuit}$\thanks{\;\;Work done during an internship at Tencent AI Lab. The work described in this paper is also partially supported by a grant from the Research Grant Council of the Hong Kong Special Administrative  Region, China (Project Code: 14200719).}\ \ \ \ \ Deng Cai$^{\heartsuit}$\thanks{\;\;Corresponding author.}\ \ \ \ \ Zhisong Zhang$^{\heartsuit}$\ \ \ \ \ Wai Lam$^{\spadesuit}$\ \ \ \ \ Shuming Shi$^{\heartsuit}$\\
   $^\heartsuit$Tencent AI Lab\ \ \ \ \ $^\spadesuit$The Chinese University of Hong Kong\\
   \texttt{ \{wwxu,wlam\}@se.cuhk.edu.hk} \\
\texttt{ \{jcykcai,zhisonzhang,shumingshi\}@tencent.com}
}

\usepackage{cleveref}
\crefname{section}{§}{§§}
\Crefname{section}{§}{§§}
\begin{document}

\maketitle

\begin{abstract}
As humans, we consistently interact with our peers and receive feedback in the form of natural language. This language feedback allows us to maintain appropriate behavior, and rectify potential errors. The question arises naturally: can we use language feedback to align large language models (LLMs)? In contrast to previous research that aligns LLMs with scalar rewards, we present the first systematic exploration of alignment through the lens of language feedback (i.e., judgment).
We start with an in-depth investigation of potential methods that can be adapted for aligning LLMs with judgments, revealing that these methods cannot fully capitalize on judgments. To facilitate more effective utilization of judgments, we propose a novel framework, Contrastive Unlikelihood Training (CUT), that allows for fine-grained inappropriate content detection and correction based on judgments.
Our results show that, with merely 1317 off-the-shelf judgment data, CUT can beat the 175B DaVinci003 and surpass the best baseline by 50.84 points on AlpacaEval using LLaMA2-13b. CUT can also align LLMs in an iterative fashion using up-to-date model-specific judgments, improving performance from 81.09 to 91.68 points on AlpacaEval using LLaMA2-chat-13b. Further analysis suggests that judgments hold greater potential in LLM alignment than rewards.\footnote{\;\;Code available at: \url{https://github.com/wwxu21/CUT}}
\end{abstract}
\section{Introduction}
Large language models (LLMs) acquire substantial world knowledge and reasoning capabilities through large-scale pre-training \citep{brown2020language,du-etal-2022-glm,touvron2023llama}. To unleash the power of pre-trained LLMs for real-world applications, it is crucial to ensure that LLMs can follow human preferences and values \citep{ouyang2022training}. This process, known as alignment, is critical for making artificial intelligence a helpful and reliable ally for humanity \citep{wang2023aligning}.

\begin{figure*}
    \centering
    \includegraphics[width=0.95\textwidth]{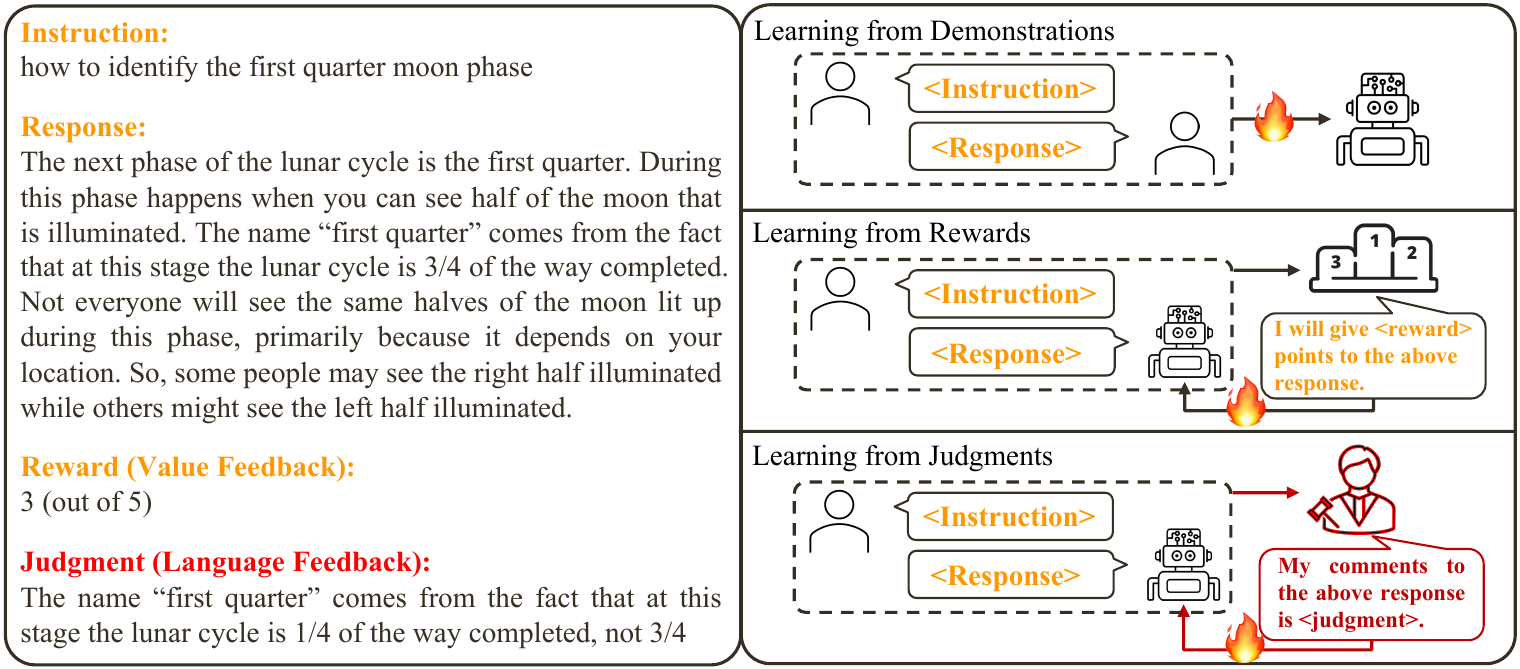}
    \caption{The illustration of three paradigms for aligning LLMs.}
    \label{fig:intro}
    \vspace{-10pt}
\end{figure*}

Figure \ref{fig:intro} illustrates three paradigms to achieve alignment. The most straightforward one is \textit{learning from demonstrations}, wherein demonstrations of desired responses to a set of instructions are collected and used to fine-tune LLMs in a supervised fashion \citep{wei2022finetuned,ouyang2022training}. However, the performance gains diminish rapidly as the data size increases \citep{zhou2023lima,fu-etal-2024-disperse}. In contrast, \textit{learning from feedback} (rewards or judgements) offers a more scalable approach \citep{ouyang2022training,bai2022training}. One significant advantage of feedback over demonstrations is that feedback can convey both positive and negative aspects, enabling the model to discern desirable and undesirable outcomes. In addition, feedback is tailored to the current model, adhering to the principle of teaching according to the learner's aptitude.

Prior research on learning from feedback primarily focuses on value feedback (i.e., scalar rewards), employing reinforcement learning (RL) algorithms, such as PPO \citep{schulman2017proximal}, to optimize an LLM to maximize the rewards of its generated responses. Nevertheless, scalar rewards are information-sparse for solely indicating the goodness of a response. On the other hand, language feedback (i.e., judgment) can offer more nuanced commendations and critiques through natural language expressions \citep{saunders2022self}. Specifically, judgments can elucidate the specific aspects that are good or bad, the rationale behind their evaluation, and suggestions for improvement. The above suggests that aligning LLMs with judgments can be more advantageous.

In this study, we present an extensive investigation of potential methods that can be adapted for \textit{aligning LLMs with judgments}. To facilitate a comprehensive aligning process, we propose a novel framework, Contrastive Unlikelihood Training (CUT), that enables fine-grained inappropriate content detection and correction based on judgments. CUT detects inappropriate content in a response by contrasting its generation probabilities under aligned and misaligned conditions and further penalizes the inappropriate content with unlikelihood training \cite{Welleck2020Neural}.

We carry out experiments for both offline and online alignment, wherein the target LLM learns from the off-the-shelf judgments and the judgments derived from self-generated responses, respectively. Extensive results on offline alignment demonstrate the effectiveness of CUT in learning from judgments in both cold-start (using unaligned base LLMs such as LLaMA2) and warm-start (using aligned base LLMs such as LLaMA2-chat) scenarios. Notably, when trained with only 1317 offline judgment data, CUT attains a winning rate of 61.06 and outperforms the best baseline by 50.84 points on AlpacaEval using LLaMA2-13b. Furthermore, our online alignment experiments show that CUT is capable of iteratively refining LLMs using model-specific judgments, with a steady performance improvement from 81.09 to 91.68 points on AlpacaEval using LLaMA2-chat-13b. Our analysis comparing rewards and judgments suggests that aligning LLMs with judgments offers significant potential and warrants future research.

Our contributions can be summarized as follows:
1) We present the first systematic exploration of aligning LLMs with judgments.
2) We introduce a novel framework, CUT, that facilitates the alignment of LLMs through fine-grained inappropriate content detection and correction based on judgments.
3) Our results showcase the effectiveness of CUT in aligning LLMs across cold-start and warm-start scenarios, generalist and specialist applications, as well as offline and online settings.
4) Our analysis indicates that judgments hold greater potential over rewards for aligning LLMs.

\section{Related Work}
\label{sec:related_work}


Existing approaches for learning from feedback can be classified into two distinct categories: prompting and fine-tuning, differentiated by whether updates to the LLMs' parameters are absent or present.

\vspace{4pt}
\noindent \textbf{Prompting.}
Prompting does not alter the parameters of LLMs. Instead, it leverages judgments on previous responses to elicit improved responses from LLMs \citep{welleck2022generating,akyurek-etal-2023-rl4f}. Judgments can be sourced from diverse aspects \citep{nathani2023maf} and the refinement process can be iterated multiple times \citep{yang2022re3,peng2023check,madaan2023self}. However, these methods rely on the in-context learning capabilities of the LLMs and consume more computation than single-pass generation \citep{brown2020language,liu2023pre}.

\vspace{4pt}
\noindent \textbf{Fine-tuning.}
Fine-tuning aims to train an LLM that can generate better responses immediately. Scalar rewards have been extensively used through the lens of RL, particularly PPO \citep{schulman2017proximal,ziegler2019fine,yang2023rlcd}. However, PPO is known to be complex and unstable \citep{zheng2023secrets}, which has attracted numerous efforts to simplify or stabilize the training process \citep{ramamurthy2023is,peng2023stabilizing,dong2023raft,touvron2023llama,rafailov2023direct,yuan2023rrhf,song2023preference,hong2024reference}. Another strand of work, named Hindsight \citep{zhang2023wisdom,liu2023languages}, transforms scalar rewards to language instructions and teach LLMs to generate responses of different qualities. There are also attempts to leverage the results of prompting for training a better model. That is, the improved response elicited by the judgment is employed as new training data \citep{scheurer2022training,scheurer2023training,yu2023constructive}. However, these methods still suffer from the incapability to learn from mistakes, which is the core spirit of learning from feedback.

\section{Preliminaries}
\label{sec:preliminaries}
In this section, we first lay out a formal problem definition of \textit{aligning LLMs with judgments} and then present a survey of three potential methods that can be adapted for tackling this problem.

\subsection{Problem Setting}
\label{sec:setting}
Suppose that there is a set of instruction-response-judgment triplets $(\boldsymbol{x}, \boldsymbol{y}, \boldsymbol{j})$, where the instruction $\boldsymbol{x}=[x_1,\ldots,x_M]$, the response $\boldsymbol{y}=[y_1,\ldots,y_N]$, and the judgment $\boldsymbol{j}=[j_1,\ldots,j_Q]$ are token sequences of length $M$, $N$, and $Q$, respectively. The response may exhibit flaws or be considered entirely satisfactory. The judgment provides an analysis of the strengths and weaknesses of the response, which can be drafted either by humans or AI models \citep{akyurek-etal-2023-rl4f,li2023generative}. The goal of aligning LLMs with judgments is to enable LLMs to retain appropriate behaviors mentioned in the strengths, and more importantly, address the weaknesses to prevent future misbehavior.

Depending on whether the responses $\boldsymbol{y}$ are from the LLM to be aligned, the learning process can be classified into two distinct types: \textit{offline alignment} and \textit{online alignment}. In offline alignment, the target LLM learns from an off-the-shelf, model-agnostic dataset. Conversely, in online alignment, the target LLM reflects on its own outputs through direct interactions with a judge. This online alignment process can be conducted iteratively, akin to how humans continuously improve their skills by receiving ongoing feedback from others over time.

\begin{table*}[t]
    \centering
    \small
    \setlength{\tabcolsep}{1mm}{
    \begin{tabular}{@{}l|c|c|c|c|c@{}}
        \toprule
        
       &\textbf{Instruction}: $\boldsymbol{x}$  & \textbf{Response}: $\boldsymbol{y}$ &  \textbf{Judgment}: $\boldsymbol{j}$ & $\boldsymbol{x}\xrightarrow{}\boldsymbol{y}$   & $[\boldsymbol{x},\boldsymbol{j}]\xrightarrow{}\boldsymbol{y}$\\
        \midrule
        \multirow{3}{*}{\rotatebox{90}{Align-P}} & \begin{tabular}[t]{@{}p{49mm}@{}}James buys 5 packs of beef that are 4 pounds each.  The price of beef is \$5.50 per pound.  How much did he pay? \end{tabular}  & \begin{tabular}[t]{@{}p{41mm}@{}} He bought 5 * 4 = 20 pounds of beef. So he paid 20 * 5.5 = \red{\$110}. \end{tabular} & \begin{tabular}[t]{@{}p{32mm}@{}} Your response to the instruction is satisfactory. \end{tabular} &\usym{2713} &\usym{2713}\\ \midrule
        \multirow{3}{*}{\rotatebox{90}{Align-N}} &\begin{tabular}[t]{@{}p{49mm}@{}}James buys 5 packs of beef that are 4 pounds each.  The price of beef is \$5.50 per pound.  How much did he pay? \end{tabular} & \begin{tabular}[t]{@{}p{41mm}@{}} Each pack was 5 pounds and it cost 5.50. So 5 * 5.50 = \blue{\$27.50}.\end{tabular}  & \begin{tabular}[t]{@{}p{32mm}@{}} The answer forgets to multiply the total amount of pounds of beef (5*4).\end{tabular} &\usym{2717} &\usym{2713}\\ \midrule
        \multirow{3}{*}{\rotatebox{90}{Misalign}} &\begin{tabular}[t]{@{}p{49mm}@{}}James buys 5 packs of beef that are 4 pounds each.  The price of beef is \$5.50 per pound.  How much did he pay? \end{tabular} & \begin{tabular}[t]{@{}p{41mm}@{}} Each pack was 5 pounds and it cost 5.50. So 5 * 5.50 = \blue{\$27.50}.\end{tabular}  & \begin{tabular}[t]{@{}p{32mm}@{}} \orange{Your response to the instruction is satisfactory.}\end{tabular} &\usym{2717} &\usym{2717}\\ \bottomrule
    \end{tabular}}
    \vspace{-4pt}
        \caption{The illustration of three categories of alignment data. $\boldsymbol{x}\xrightarrow{}\boldsymbol{y}$   and $[\boldsymbol{x},\boldsymbol{j}]\xrightarrow{}\boldsymbol{y}$ indicate if the response aligns with the instruction or the combination of instruction and judgment, respectively.}
    \label{tab:data}
    \vspace{-12pt}
\end{table*}

\subsection{Potential Solutions}
\label{sec:solutions}
\noindent \textbf{Forward Prediction} refers to sequentially predicting the response and its judgment \citep{chen2024gaining}, which was originally proposed in dialogue generation \citep{weston2016dialog,li2016dialogue}. It can be seamlessly adapted to our problem.
Specifically, the LLM is trained with the maximum likelihood estimation (MLE) objective to first generate the response $\boldsymbol{y}$ based on the instruction  $\boldsymbol{x}$ and subsequently generate the judgment  $\boldsymbol{j}$ based on the combined sequence  $[\boldsymbol{x},\boldsymbol{y}]$.
\begin{equation}
\label{eq:fp}
\small
        L_{f}=-\frac{1}{N}\sum_t\log p(y_t|y_{<t},\boldsymbol{x})-\frac{1}{Q}\sum_t\log p(j_t|j_{<t},\boldsymbol{y},\boldsymbol{x})
\end{equation}

\noindent \textbf{Imitation learning from language feedback} (ILF) asks the LLM to refine the initial response $\boldsymbol{y}$ based on the feedback $\boldsymbol{j}$  to be an improved response $\hat{\boldsymbol{y}}$. 
\begin{equation}
\small
\hat{\boldsymbol{y}} = \textbf{LLM}(\boldsymbol{x}, \boldsymbol{y}, \boldsymbol{j})
\end{equation}
\begin{itemize}[leftmargin=*,topsep=4pt]
\setlength{\itemsep}{0pt}
\setlength{\parskip}{0pt}
\setlength{\parsep}{0pt}
    \item \textbf{ILF-MLE:} The improved response $\hat{\boldsymbol{y}}$ can be directly paired with the initial instruction $\boldsymbol{x}$ to fine-tune the LLM under the MLE objective \citep{bai2022constitutional,scheurer2022training,scheurer2023training}.
\begin{equation}
\small
\label{eq:ilf}
L_{i}^{mle} = -\frac{1}{N}\sum_t\log p(\hat{y}_t|\hat{y}_{<t}, \boldsymbol{x})
\end{equation}
\item \textbf{ILF-DPO:} 
\citet{yu2023constructive} demonstrate that the improved response $\hat{\boldsymbol{y}}$ and the original response $\boldsymbol{y}$ can be used jointly as a pairwise comparison, where $\hat{\boldsymbol{y}}$ is a more preferred response to $\boldsymbol{x}$ compared to $\boldsymbol{y}$. As a result, preference learning algorithms, such as direct preference optimization (DPO) \citep{rafailov2023direct}, can be adopted to fine-tune the LLM: $L_{i}^{dpo}=\textbf{DPO}(\boldsymbol{x}, \boldsymbol{y}, \hat{\boldsymbol{y}})$.
\end{itemize}

\noindent \textbf{Hindsight}  rewrites the instruction $\boldsymbol{x}$ based on the scalar rewards received by the response $\boldsymbol{y}$ \citep{zhang2023wisdom,liu2023languages}. For instance, if a response receives a scalar reward below a certain threshold, the phrase ``generate a good answer" is appended to the original instruction. This approach can be naturally extended to our problem setting. Concretely, the LLM is trained to generate the response $\boldsymbol{y}$ conditioned on the sequence $[\boldsymbol{x}, \boldsymbol{j}]$.
\begin{equation}
\small
\label{eq:hindsight}
    L_{h}=-\frac{1}{N}\sum_t\log p(y_t|y_{<t}, \boldsymbol{x}, \boldsymbol{j})
\end{equation}

However, in Forward Prediction, learning to generate judgments does not necessarily translate into enhanced response generation, given that response generation precedes judgment generation. The indirect usage of judgment in ILF limits its capacity to spot and rectify weaknesses underscored in judgments. Hindsight employs unsatisfactory responses as MLE targets, which inevitably increases the risk of generating unsatisfactory responses. In summary, we contend that existing methods cannot fully capitalize on judgments, which motivates us to design a better solution.

\section{Contrastive Unlikelihood Training}
To overcome the limitations mentioned in \cref{sec:preliminaries}, we propose CUT, a fine-tuning framework to align LLMs with judgments. The core idea of CUT is summarized as \textbf{Learning from Contrasting}. We contrast the response generation under different conditions to shed light on the appropriate behavior that the LLM should keep, as well as the specific content necessitating adjustments. Based on these insights, we use MLE for appropriate content and UT  \citep{Welleck2020Neural} for inappropriate content.

\subsection{Incorporating Judgments for Alignment}
We call an instruction-response pair ``aligned" if the response follows the instruction faithfully and satisfies human expectations $\boldsymbol{x}\xrightarrow{}\boldsymbol{y}$.
Otherwise, a judgment describes the errors or deficiencies present in the response.
Assuming the task is to generate a response that intentionally fulfills the judgment, it can be inferred that the response always aligns with the combined input of instruction and judgment $[\boldsymbol{x},\boldsymbol{j}]\xrightarrow{}\boldsymbol{y}$. Based on the idea, we construct three types of alignment data, depicted in Table \ref{tab:data}.

\vspace{2pt}
\noindent \textbf{Align-P:}
The LLM produces a satisfactory response $\boldsymbol{y}$ to the instruction $\boldsymbol{x}$. Therefore, a positive judgment $\boldsymbol{j}$ is conferred to praise the commendable performance. The response $\boldsymbol{y}$ is aligned with the instruction $\boldsymbol{x}$ as well as the combined input $[\boldsymbol{x}, \boldsymbol{j}]$.

\vspace{2pt}
\noindent \textbf{Align-N:}
The LLM makes some mistakes in its generation, resulting in an unsatisfactory response $\boldsymbol{y}$. Consequently, a negative judgment $\boldsymbol{j}$ details the corresponding critiques. For Align-N, $\boldsymbol{y}$ is not aligned with original instruction $\boldsymbol{x}$. However, when considering $\boldsymbol{x}$ and $\boldsymbol{j}$ as a whole, $\boldsymbol{y}$ is indeed aligned with the combined input $[\boldsymbol{x}, \boldsymbol{j}]$.

\vspace{2pt}
\noindent \textbf{Misalign:}
The authentic negative judgment in Align-N is substituted with a fake positive judgment $\boldsymbol{j}$.  In this case, the response $\boldsymbol{y}$ is not aligned with either the original instruction $\boldsymbol{x}$ or the combination of instruction and judgment $[\boldsymbol{x}, \boldsymbol{j}]$.

\subsection{Learning from Contrasting}
\label{sec:contrast}
With the above three categories of alignment data. We can deduce two notable contrasts that provide valuable insights to guide the alignment of LLMs.

\begin{figure}
    \centering
    \includegraphics[width=0.45\textwidth]{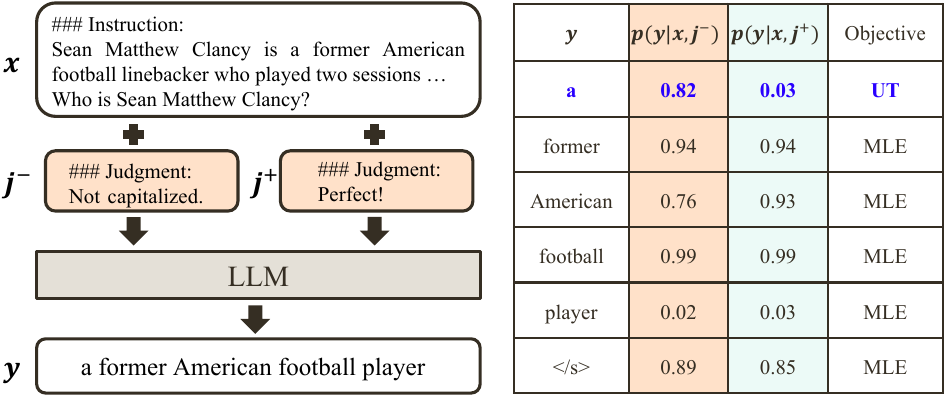}
    \caption{Generation probability of identical output text under Align-N (left) and Misalign (right) contexts.}
    \label{fig:toy}
    \vspace{-10pt}
\end{figure}

\vspace{4pt}
\noindent \textbf{Align-N vs. Misalign:} 
The major difference between these two is that they show opposite polarities in the task of $[\boldsymbol{x},\boldsymbol{j}]\xrightarrow{}\boldsymbol{y}$. Thanks to the strong in-context learning capabilities of LLMs, the alignment flip from Align-N (aligned) to Misalign (misaligned) is often accompanied by decreased generation probabilities of the response, particularly for tokens that exhibit a strong correlation with the authentic negative judgment. Figure~\ref{fig:toy} presents a simple example, wherein the response commits a minor capitalization issue. The LLM assigns a considerably higher probability for ``a" when taking the authentic negative judgment $\boldsymbol{j}^-$ instead of the fake positive judgment $\boldsymbol{j}^+$ as additional input, precisely at the point where the LLM commits the error.

To take advantage of the above contrast, we feed Align-N and Misalign examples to the LLM to get token generation probabilities $p(y_t|\boldsymbol{y}_{<t}, \boldsymbol{x}, \boldsymbol{j}^-)$ and $p(y_t|\boldsymbol{y}_{<t}, \boldsymbol{x}, \boldsymbol{j}^+)$ separately. We consider the tokens that display a substantially increased generation probability when conditioned on $\boldsymbol{j}^-$ compared to $\boldsymbol{j}^+$ as inappropriate tokens (e.g., ``a'' in Figure~\ref{fig:toy}).
Concretely, the following criterion is adopted:
\begin{equation}
\label{eq:identification}
\small
\begin{aligned}
U = &\{t \;| \; p(y_t | \boldsymbol{y}_{<t}, \boldsymbol{x}, \boldsymbol{j}^-) - \lambda \cdot p(y_t|\boldsymbol{y}_{<t}, \boldsymbol{x},\boldsymbol{j}^+) > 0\}
\end{aligned}
\end{equation}
where $\lambda$ is a hyperparameter to tradeoff the precision and recall of detecting inappropriate tokens.

We apply the UT on the identified inappropriate tokens for pushing the LLM to explore alternative generations. Motivated by the focal loss \citep{lin2017focal}, we introduce a dynamic weighting mechanism. This mechanism is designed to modulate the penalty applied to inappropriate tokens in proportion to their degree of inappropriateness. For other tokens, we use the standard MLE loss:
\begin{equation}
\label{eq:CUT}
\small
\begin{aligned}
    L_1=&-\frac{1}{N}(\sum_{t\notin U}\log p(y_t|\boldsymbol{y}_{<t}, \boldsymbol{x}) \\
    &+ \sum_{t\in U}\alpha p(y_t | \boldsymbol{y}_{<t}, \boldsymbol{x}, \boldsymbol{j}^-)^{\gamma}\log (1- p(y_t|\boldsymbol{y}_{<t}, \boldsymbol{x})))    
\end{aligned}
\end{equation}
where $\alpha p(y_t | \boldsymbol{y}_{<t}, \boldsymbol{x}, \boldsymbol{j}^-)^{\gamma}$ is the dynamic weight term. $\alpha$ and $\gamma$ are two hyper-parameters. A higher value of $p(y_t | \boldsymbol{y}_{<t}, \boldsymbol{x}, \boldsymbol{j}^-)$ suggests that the response tokens have a stronger correlation with negative judgments. Consequently, such tokens are more prone to be inappropriate and are thus subjected to a larger unlikelihood penalty.

\vspace{4pt}
\noindent \textbf{Align-P vs. Align-N:}
Despite both Align-P and Align-N are aligned in terms of $[\boldsymbol{x},\boldsymbol{j}]\xrightarrow{}\boldsymbol{y}$, only Align-P is aligned when solely considering the instruction ($\boldsymbol{x}\xrightarrow{}\boldsymbol{y}$). Essentially, it suggests that the LLM should output different responses depending on whether a negative judgment is incorporated or not. Therefore, the comparison provides valuable information for the LLM to discern satisfactory and unsatisfactory responses. Specifically, we train on this comparison with the following MLE objective:
\begin{align}
\small
\begin{aligned}
L_2 =&-\frac{\mathbbm{1}(\boldsymbol{x}\xrightarrow{}\boldsymbol{y})}{N}\sum_t\log p(y_t|\boldsymbol{y}_{<t}, \boldsymbol{x})\\
&-\frac{(1 - \mathbbm{1}(\boldsymbol{x}\xrightarrow{}\boldsymbol{y}))}{N}\sum_t\log p(y_t|\boldsymbol{y}_{<t}, \boldsymbol{j},\boldsymbol{x})
\end{aligned}
\end{align}
where $\mathbbm{1}(\boldsymbol{x}\xrightarrow{}\boldsymbol{y})$ is an indicator function that returns $1$ if $x$ and $y$ are aligned, and $0$ otherwise.

Finally, the overall loss of CUT combines the losses from the two contrasts: $L_{\text{CUT}} = L_1 + L_2$.


\subsection{Relation to Prior Solutions}
We discuss the connections of CUT to prior solutions of learning from judgments.

\noindent \textbf{Forward Prediction} hopes that the judgment generation could indirectly boost its response generation abilities. In contrast, CUT directly utilizes judgments to teach the LLM how to generate satisfactory responses and avoid unsatisfactory ones.

\noindent \textbf{ILF} assumes judgments can always elicit improved responses. This solution essentially learns from such pseudo improved response. Conversely, CUT can directly learn from misaligned data.

\noindent \textbf{Hindsight} learns to generate responses of different qualities at the risk of increasing the likelihood of unsatisfactory responses. In comparison to Hindsight, CUT mitigates this issue by
incorporating both likelihood and unlikelihood training objectives.

\section{Experiments}

\begin{table*}[t]
    \centering
    \small
    \setlength{\tabcolsep}{2mm}{
    \begin{tabular}{@{}llc|ccccc|cccc@{}}
    \toprule 
    \multicolumn{2}{l}{\textbf{Method}}  & \textbf{Objective}     & \textbf{ARC} & \textbf{HellaSwag} & \textbf{MMLU}  & \textbf{TruthfulQA} & \textbf{Avg.} & \textbf{AlpacaEval}\\
    \midrule
    \multirow{7}{*}{\rotatebox{90}{LLaMA2-13b}} &Base   & - & 59.72 & 81.39 & 54.97 & 36.28 &58.09& 1.87\\
    &Forward Prediction                         & MLE&56.91&81.03&54.35& 34.28 & 56.64  & 7.11\\
    &Hindsight                       & MLE&58.11&81.33&55.33& 35.61 & 57.60& 10.22\\
    &ILF-MLE                  & MLE& 58.36 & 81.15 & 53.76 & 37.03 &57.58& 4.01\\
    & ILF-DPO & DPO & 58.79 & 81.07 & 55.48 & 41.84 & 59.3 &3.11 \\
        \cmidrule{2-9}
     &CUT (ours)                      & MLE+UT &\textbf{60.84}&\textbf{81.44}&\textbf{55.78}& \textbf{49.33} & \textbf{61.85} &  \textbf{61.06}\\
    \midrule  \midrule
    \multirow{7}{*}{\rotatebox{90}{LLaMA2-chat-13b}} &Base & -&58.02 & 79.89 & 54.52 &45.44 & 59.47 &81.09\\
    &Forward Prediction                         & MLE& 52.22 & 78.16 & 53.06&37.69 & 55.28 & 33.21\\
    &Hindsight                       & MLE& 53.92 & 78.58 & 54.15 & 39.01 & 56.42& 36.67\\
    &ILF-MLE            & MLE & 58.36 & \textbf{81.15} & 53.76 & 45.65 & 59.73 & 79.31\\
    & ILF-DPO & DPO & \textbf{58.81} &	80.04&	54.98	&51.51	& 61.34 &83.22& \\
    \cmidrule{2-9}
    &CUT (ours)                    & MLE+UT&58.45& 79.86&\textbf{55.00}& \textbf{52.58} & \textbf{61.47}&  \textbf{90.73}\\
    \bottomrule
\end{tabular}}
\caption{Results on General Instruction-following. Objective column denotes the fine-tuning objective.}
\label{tab:13B_shepherd}
\vspace{-10pt}
\end{table*}

\begin{table}[t]
    \centering
    \small
    \setlength{\tabcolsep}{0.5mm}{
    \begin{tabular}{@{}ll|ccccc}
    \toprule 
    \multicolumn{2}{l|}{\textbf{Model}} & \textbf{rouge1} & \textbf{rouge2} & \textbf{rougeL} & \textbf{rougeLsum}  \\

    \midrule
    \multirow{7}{*}{\rotatebox{90}{LLaMA2-13b}}&Base              & 12.91 & 6.33 & 10.10 & 10.87 \\
    &Forward Prediction   & 42.42 & 28.02 & 38.45 & 38.51 \\
    &Hindsight           & 38.33 & 25.49 & 35.26 & 35.29\\
    &ILF-MLE                   & 28.51 & 16.68 & 25.36 & 25.44 \\
    &ILF-DPO                 & 11.31 & 7.45 & 10.23 & 10.77 \\
    \cmidrule{2-6}
    &CUT (ours)         & \textbf{45.39} & \textbf{28.40} & \textbf{39.84} & \textbf{39.89} \\
        \midrule \midrule 
    \multirow{7}{*}{\rotatebox{90}{LLaMA2-chat-13b}} & Base          &29.21 & 15.00 & 22.78 & 23.44\\
    &Forward Prediction  & 42.44 & 28.12 & 38.48 & 38.46\\
    &Hindsight           & 41.02 & 27.48 & 37.42 & 37.46 \\
    &ILF-MLE                  &39.21 & 27.93 & 34.35 & 34.66\\
    &ILF-DPO                  & 33.90 & 19.81 & 28.01 & 28.18\\
    \cmidrule{2-6}
    &CUT (ours)         & \textbf{45.31} & \textbf{29.04} & \textbf{39.96} & \textbf{40.12}\\
    \bottomrule
\end{tabular}}
\caption{Results on the summarization task.}
\label{tab:13B_summary}
\vspace{-10pt}
\end{table}

To provide a comprehensive assessment of CUT, we implement it in two alignment scenarios: offline alignment and online alignment. In the offline alignment experiments, we perform extensive analysis on the adaptability and universality of CUT across different model and task configurations. In the online alignment experiments, we additionally explore the possibility of building an automatic judgment model. Lastly, to highlight the potential of aligning LLMs with judgments, we establish a comparison between learning from rewards and learning from judgments.

\noindent \textbf{Tasks.}
We experiment on both general instruction-following and a specific NLP task (summarization).
For \textbf{Instruction following}, we evaluate models on both AlpacaEval and four additional conventional NLP benchmarks: 25-shot ARC, 10-shot HellaSwag, 5-shot MMLU, and 0-shot TruthfulQA. For AlpacaEval, we report the winning rate of the responses generated by our models against DaVinci003 using GPT4 as the judge. The four conventional NLP benchmarks are ranking-based and we report accuracies.
For \textbf{Summarization}, we use the dataset from \citet{saunders2022self} and report ROUGE scores \citep{lin-2004-rouge} on 1939 test examples. See Appendix \ref{sec:tasks} for more details.

\noindent \textbf{Baselines.}
The baselines include the base model without further fine-tuning, and the three groups of judgment-based alignment methods:
(1) The Forward Prediction method described in Eq. \ref{eq:fp} \citep{weston2016dialog}; 
(2) The Hindsight method described in Eq. \ref{eq:hindsight} \citep{zhang2023wisdom};
(3)  ILF-MLE described in Eq. \ref{eq:ilf} \citep{scheurer2022training}, and ILF-DPO \citep{yu2023constructive} that change the learning objective from MLE to DPO.
The details of the model implementations are provided in Appendix \ref{sec:impl}.

\subsection{Offline Alignment}
\label{sec:offline}
The offline setting utilizes off-the-shelf instruction-response-judgment triplets for alignment. This aims to validate and analyze CUT in controlled environments prior to initiating the costly process of model-specific judgment annotation. For instruction following, we train models with 1317 examples from Shepherd \citep{wang2023shepherd}. For summarization, we use the 10827 training examples with judgment annotations from \citet{saunders2022self}.

\noindent \textbf{Results.}
The results of the general instruction-following and summarization are presented in Table \ref{tab:13B_shepherd} and \ref{tab:13B_summary}, respectively. For cold-start scenarios (LLaMA2-13b as the base model), CUT improves the winning rate on AlpacaEval from 1.87 to 61.06, where CUT beats the 175B DaVinci003 and surpasses the best baseline (Hindsight) by 50.84 points.
Moreover, CUT improves the base model by 13.05 points on TruthfulQA. This implies that CUT can effectively mitigate hallucinations. Conversely, most baselines improve marginally or experience performance drops on TruthfulQA. This is likely due to their application of the MLE objective on error-prone responses, which reduces factuality in response generation.
In terms of ARC, HellaSwag, and MMLU, CUT remains competitive with the base model, indicating CUT suffers less from the alignment tax problem \citep{ouyang2022training}. For single NLP task (i.e., summarization) experiments, CUT surpasses the best baseline (i.e., Forward Prediction) by 1.38 rougeLsum scores. Overall, the results show that CUT is effective in transforming LLMs into both performant generalist and specialist models. 

The performance superiority of CUT in warm-start scenarios (LLaMA2-chat-13b as the base model) are consistent with the cold-start ones. The two ILF methods (ILF-MLE and ILF-DPO) outperform methods from Forward Prediction and Hindsight groups on AlpacaEval in warm-start but perform worse in cold-start scenarios. This may be due to that ILF methods heavily rely on the base model in producing high-quality improved responses, making it less effective in cold-start scenarios.

\begin{table}
    \centering
    \small
    \setlength{\tabcolsep}{1mm}{
    \begin{tabular}{@{}l|cc}
    \toprule 
    \textbf{Model}    & \textbf{Generalist} &\textbf{Specialist}\\
    \midrule
    LLaMA2-chat-13b                     &45.44 & 23.44\\
    \midrule
    CUT                      &52.58 & 40.12\\
    ~- $L_1$               &39.01   & 37.46\\
    ~- first part of $L_2$ & - & 26.37\\
    ~- second part of $L_2$ & 47.24 & 33.92\\
    ~- Inappropriate Token Detection             &0   & 0\\
    ~- Dynamic Weighting               &48.84   & 40.05\\
    \bottomrule
\end{tabular}}
\caption{Effect of CUT designs. We report the results on TruthfulQA (Acc.) and summarization test set (rougeLsum) for general instruction-following (Generalist) and Summarization (Specialist) respectively. ``-'' indicates no Align-P examples in the Generalist training set.}
\label{tab:ablation}
\vspace{-15pt}
\end{table}

\vspace{4pt}
\noindent \textbf{Ablation Study.}
To investigate the effectiveness of two contrasts employed by CUT, we perform ablation studies by eliminating certain training signals. The results are shown in Table \ref{tab:ablation}. Removing the contrast between Align-N and Misalign (- $L_1$) substantially reduces the performance of TruthfulQA. This finding highlights that the UT objective plays a crucial role in mitigating hallucinations. The exclusion of the contrast between Align-P and Align-N can be implemented in two ways. We can either remove the first part or the second part of $L_2$. As seen, the impact of removing Align-P is more pronounced than removing Align-N on the summarization task. This may be attributed to the necessity of positive examples for adapting the LLM to a specific task. Furthermore, we introduce an additional ablated variant in which the inappropriate token detection (Eq. \ref{eq:identification}) is omitted (- Inappropriate Token Detection). Concretely, we simply apply UT for all tokens in misaligned responses instead. Intriguingly, we find that this approach fails to converge during training. This observation underscores the importance of inappropriate token detection. Lastly, removing the dynamic weighting term ($p(y_t | \boldsymbol{y}_{<t}, \boldsymbol{x}, \boldsymbol{j}^-)^{\gamma}$ in Eq. \ref{eq:CUT}) also impacts the effectiveness of CUT, particularly in general instruction-following tasks.
\begin{table}
    \centering
    \small
    \setlength{\tabcolsep}{0.5mm}{
    \begin{tabular}{@{}l|ccccc@{}}
    \toprule 
    \textbf{Method}    & \textbf{ARC} & \textbf{HeSw} & \textbf{MMLU}  & \textbf{TQA} & \textbf{AlpacaEval}\\
    \midrule \multicolumn{6}{c}{ \em \small Different Model Size} \\\midrule
    LLaMA2-7b-chat &	51.45	&78.63 &	43.60&	43.71&	71.40 \\
    + CUT &\textbf{53.16}& \textbf{79.23}&\textbf{46.95}& \textbf{51.40} & \textbf{86.94}\\ \midrule
    LLaMA2-13b-chat &58.02 & \textbf{79.89} & 54.52 &45.44 &81.09\\
    + CUT &\textbf{58.45}& 79.86&\textbf{55.00}& \textbf{52.58} &   \textbf{90.73}\\ \midrule
    LLaMA2-70b-chat &65.27&83.89 & \textbf{63.07} &53.09&92.70 \\
    + CUT &\textbf{66.30}& \textbf{84.00}&62.71& \textbf{55.45} & \textbf{93.04}\\ 
    \midrule \multicolumn{6}{c}{ \em \small Different Backbone Models} \\\midrule
    Mistral-7b-it-v1 & 53.67 &  74.00& 54.66 &55.54  & 69.33\\
    + CUT & \textbf{54.27} & \textbf{75.70} & \textbf{54.98} & \textbf{57.61} &\textbf{82.75}\\ \midrule
    gemma-7b-it &  51.37 & 71.67&51.66& 30.35 & 78.51 \\
    + CUT & \textbf{51.96} & \textbf{72.53} & \textbf{52.28} & \textbf{30.72} &\textbf{81.96}\\\midrule
    llama3-8b-it &  62.20 & 78.83&65.81& 51.65 & 93.79\\
    + CUT &61.83 &78.91 &65.60 &51.82 &94.09\\
    \bottomrule
\end{tabular}}
\caption{Effect of CUT on different model sizes and different instruction-tuned models. HeSw denotes HellaSwag and TQA denotes TruthfulQA.}
\label{tab:more_models}
\vspace{-15pt}
\end{table}

\vspace{4pt}
\noindent \textbf{Adaptability of CUT.}
Table \ref{tab:more_models} presents the impact of CUT framework on a diverse array of models, spanning across multiple model sizes and various instruction-tuned backbone architectures. This examination enables a multifaceted understanding of CUT's effectiveness and its potential scalability across different model configurations.
The upper part of Table \ref{tab:more_models} focuses on the model sizes, which are analyzed on the LLaMA2-chat family across three distinct scales: 7B, 13B, and 70B.  CUT consistently improves the performance across all sizes of the LLaMA2-chat models. This shows that CUT could be scaled up into larger models.
Progressing beyond model sizes, the bottom part of Table \ref{tab:more_models} broadens the scope to include various instruction-tuned backbone models - Mistral-7b-instruct-v1 \citep{jiang2023mistral}, gemma-7b-it \citep{team2024gemma}, and llama3-8b-instruct\footnote{\url{https://llama.meta.com/llama3}}.  CUT consistently elevates performance across almost all evaluated tasks. This exploration extends the effectiveness of CUT beyond a single model family, shedding light on its adaptability and utility across different model architectures.

\begin{table*}[ht]
    \centering
    \small
    \setlength{\tabcolsep}{2mm}{
    \begin{tabular}{l|c|cccccccccc}
    \toprule 
     \textbf{Model}  & \textbf{\#J}   & \textbf{ARC} & \textbf{HellaSwag} & \textbf{MMLU}  & \textbf{TruthfulQA} & \textbf{AlpacaEval}\\
    \midrule

    LLaMA2-chat-13b &- &58.02 & \textbf{79.89} & 54.52 &45.44 &81.09\\
    \midrule
    CUT (ours) & 1317&\textbf{58.45}& 79.86&\textbf{55.00}& \textbf{52.58} &  \textbf{90.73}\\
    \midrule
    CUT 1+ (online iteration-1)    & 1000 &58.02 & 79.55& 54.62 &50.56  & 89.75\\
    CUT 2+ (online iteration-2)    & 1000 &57.94 &79.18 & 54.83&51.67  & 90.23\\
    CUT 3+ (online iteration-3)    & 1000 &58.11 &\textbf{79.99} & 55.00 &52.69  & 91.04\\
    CUT 4+ (online iteration-4)    & 1000 &58.36 &79.23 &55.02& 52.56 & \textbf{91.68}\\
    CUT 5+ (online iteration-5)    & 1000 &\textbf{58.45} &79.19 &\textbf{55.20} & \textbf{52.96} & 90.68\\
    \bottomrule
\end{tabular}}

\caption{The results of online iterative alignment. \#J denotes the number of judgment data used in each iteration.}
\label{tab:iter}
\vspace{-15pt}
\end{table*}

\subsection{Online Alignment}
\label{sec:iterative}
In this section, we move to a more pragmatic scenario where the target LLM directly learns from the judgments associated with its own responses. As mentioned in \cref{sec:setting}, the online alignment process can be conducted iteratively, akin to how humans continuously refine their behaviors through ongoing feedback. Specifically, we apply the following three steps repeatedly:
\begin{itemize}[leftmargin=*,topsep=4pt]
\setlength{\itemsep}{0pt}
\setlength{\parskip}{0pt}
\setlength{\parsep}{0pt}
    \item \textbf{Step 1:} Collect a set of instructions $\boldsymbol{x}$, and obtain the responses $\boldsymbol{y}$ from the target model.
    \item \textbf{Step 2:} Annotate judgments $\boldsymbol{j}$ for the responses.
    \item \textbf{Step 3:} Apply CUT to fine-tune the target model with $\{\boldsymbol{x}, \boldsymbol{y},\boldsymbol{j}\}$.
\end{itemize}
where the target LLM is LLaMA2-chat-13b. In each iteration, we sample 1000 distinct instructions from Stanford Alpaca \citep{2023TaoriAlpaca}. We ask GPT4 for drafting judgments, which has been proven to produce high-quality annotations \citep{cui2023ultrafeedback}. Annotation details are elaborated in Appendix \ref{app:annotation}. 
Note that most responses from LLaMA2-chat-13b receive positive judgments, resulting in a large proportion of Align-P examples. We found downsampling Align-P examples is beneficial to the online alignment (see Appendix \ref{sec:down}).
We evaluate models on ARC, HellaSwag, MMLU, TruthfulQA, and AlpacaEval.

\begin{figure*}
    \centering
    \includegraphics[width=1\textwidth]{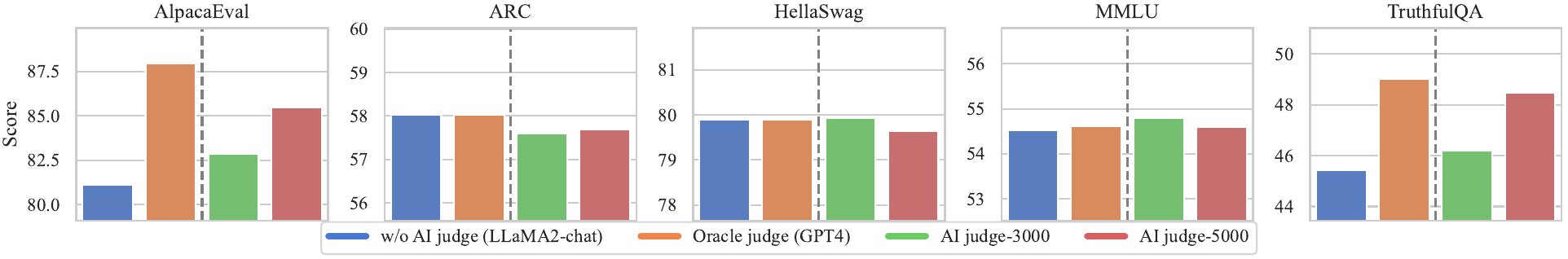}
    \vspace{-15pt}
    \caption{The results of online alignment with different AI judges.}
    \label{fig:aijudge}
    \vspace{-10pt}
\end{figure*}

\vspace{4pt}
\noindent \textbf{Results.}
Table \ref{tab:iter} shows the results of online iterative alignment. In the first iteration, online alignment exhibits superior performance over offline alignment on both TruthfulQA and AlpacaEval. This observation implies that model-specific judgments are more effective for alignment. More importantly, the alignment continues to improve with more iterations, where the performance rises from 81.09 to 91.68 on AlpacaEval after four iterations. However, the performance improvement ceases at the fifth iteration. We speculate two possible explanations for this occurrence: (1) the judgments provided by GPT-4 contain certain inaccuracies, making them insufficient to effectively align a strong target model like our CUT 4+. (2) The target model may exhibit a knowledge deficiency in specific domains, such as mathematics and science, which cannot be adequately addressed through judgments.
We also provide a case study in Appendix \ref{sec:case1}.

\subsubsection{Training A Judgment Model}
In the previous experiments, we show that CUT is effective in aligning LLMs with judgments annotated by humans or GPT4. However, human annotations can be very expensive. The use of GPT4 assumes that a very strong LLM already exists. Next, we investigate the possibilities of developing an AI judge based on the target LLM. 

\vspace{4pt}
\noindent \textbf{Setup.}
we train AI judges with different amounts of judgment data $\{3000, 5000\}$ collected in \cref{sec:iterative}.
Then, we sample 1000 new instructions from Stanford Alpaca, obtain the corresponding responses from the target model (i.e., LLaMA2-chat-13b), and label judgments with our AI judges.
These new judgment triplets are used to align the target model.

\vspace{4pt}
\noindent \textbf{Results.}
Figure~\ref{fig:aijudge} shows that AI judge-5000, trained with 5000 judgment data, is beneficial for aligning the target LLM, which leads to improvements of 3.02 and 4.17 points compared to LLaMA2-chat-13b on TruthfulQA and AlpacaEval respectively.
In contrast, AI Judge-3000, using a smaller training dataset, shows limited effectiveness. The comparison suggests that training a capable AI judge necessitates a moderate number of high-quality training instances. As a result, it is feasible to train AI judges to align the LLM. However, the quality of the AI judge remains a crucial factor in determining the success of this endeavor.

\begin{figure*}
    \centering
    \includegraphics[width=1\textwidth]{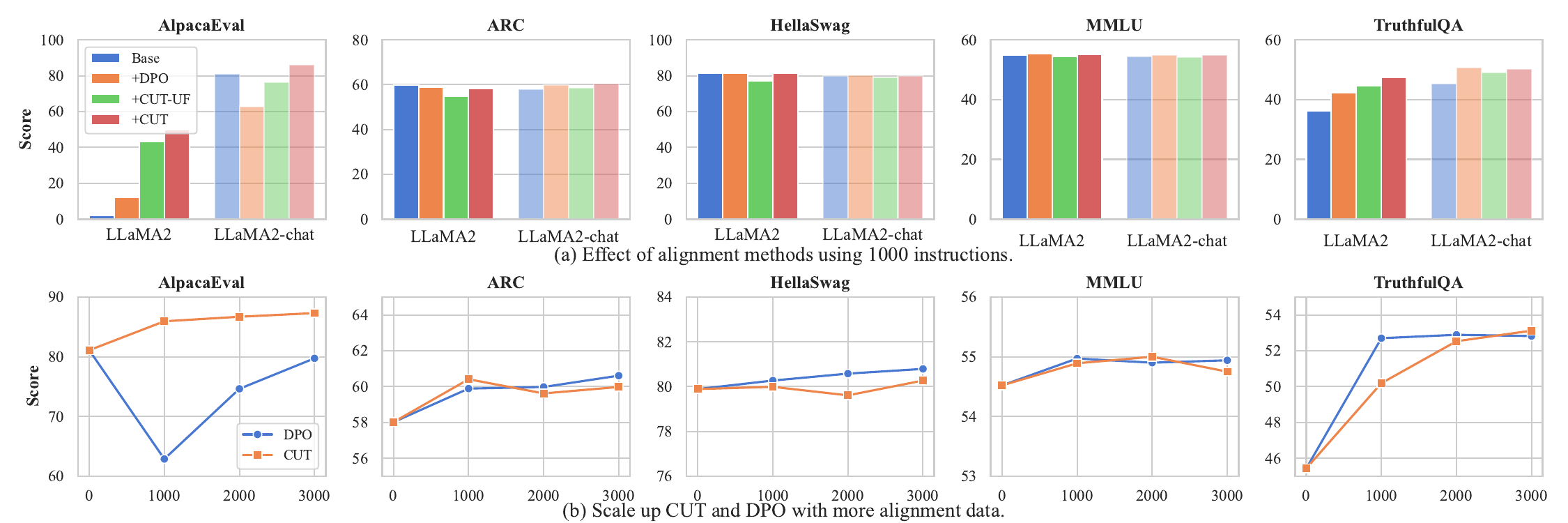}
    \vspace{-15pt}
    \caption{Comparison between reward-based DPO and judgment-based CUT.}
    \label{fig:dpo}
    \vspace{-10pt}
\end{figure*}

\subsection{Judgment vs. Reward}
Our work primarily focuses on aligning LLMs with judgments, whereas most prior research explores rewards. In this section, we aim to provide a direct comparison between these two paradigms. However, note that it is hard to conduct a fair comparison due to the distinct data formats and the potential variation in data quality.

\vspace{4pt}
\noindent \textbf{Setup.} We compare judgment-based CUT with the state-of-the-art reward-based DPO \citep{rafailov2023direct}. To maximize fairness, we leverage UltraFeedback \citep{cui2023ultrafeedback}, which contains both reward and judgment annotations produced by GPT4. Our preliminary experiments show that CUT is not good using the original judgments in UltraFeedback. We find that the reason is that the judgments in UltraFeedback tend to commend the strengths of given responses. This type of judgment is unsuitable for our CUT, as we primarily use judgments for inappropriate token detection. Therefore, we re-collect judgments on the same instruction-response pairs from GPT4 using our prompt (Appendix \ref{app:annotation}). Due to budget constraints, we randomly sample up to 3000 instructions (with 4 responses each, totaling 12,000 pairs) for annotation. 
The implementation details are as follows:
\begin{itemize}[leftmargin=*,topsep=4pt]
\setlength{\itemsep}{0pt}
\setlength{\parskip}{0pt}
\setlength{\parsep}{0pt}
    \item \textbf{DPO}:  For each of the above instructions, we formulate preference data by enumerating all possible pairs of responses from the given four, excluding pairs with the same reward value. 
    \item \textbf{CUT-UF}: We fine-tune the base model on the above instruction-response pairs and their original judgments from UltraFeedback using CUT.
     \item \textbf{CUT}: We use the same instruction-response pairs as CUT-UF but with our re-annotated judgments.
\end{itemize}

\vspace{4pt}
\noindent \textbf{Results.}
Figure \ref{fig:dpo} (a) shows the effect of three alignment methods using 1000 instructions as the alignment data.
We can observe that CUT consistently improves over CUT-UF on all five tasks for two base models, which verifies our assumption that  CUT is more effective when using critics as the judgment.
Notably, CUT surpasses DPO by a large margin of 37.54 and 23.04 points on AlpacaEval for two base models, respectively. This shows that CUT is more effective in aligning LLMs with limited alignment data (i.e., 1000 instructions).
Figure \ref{fig:dpo} (b) depicts the trends when adding more data for CUT and DPO alignment.
The performance of CUT on these tasks is generally better or comparable to that of DPO and demonstrates a positive correlation with the size of the training data provided.
The above observations suggest that judgments hold greater potential than rewards in aligning LLMs.
CUT is slightly worse than DPO on ARC, and HellaSwag. We hypothesize that the performance discrepancy is partly caused by the evaluation protocols: the four tasks are ranking-based. As suggested \citet{bansal2023peering},  methods such as DPO, which leverage ranking data in the alignment possess inherent advantages in ranking-based tasks.
We also provide a case study in Appendix \ref{sec:case2}.

\section{Conclusion}
We systematically explored the alignment of LLMs through the lens of judgments. We investigated three potential methods that can be adapted for aligning LLMs with judgments but found them unable to fully capitalize on judgments. We proposed a novel framework CUT, that enables direct and explicit learning from judgments and facilitates fine-grained inappropriate content detection and correction. Extensive evaluations demonstrated the effectiveness of our CUT in various settings, including offline and online, specialist and generalist, as well as cold-start and warm-start scenarios. For example, the online alignment experiments showed that CUT can iteratively improve LLMs with up-to-date model-specific judgments, akin to how humans progressively refine their behaviors through ongoing feedback. Our analysis comparing rewards and judgments suggested that aligning LLMs with judgments is a promising research area.

\section*{Limitations}
\paragraph{Quality of Judgment Models}
Despite the positive alignment results of our AI judge mentioned in Figure~\ref{fig:aijudge}, we find the quality of its generated judgments is not satisfactory and significantly inferior to those generated by GPT4.
Therefore, we discuss from the point of judgment generation and identify two limitations when interacting with AI judges:
\begin{itemize}[leftmargin=*,topsep=4pt]
\setlength{\itemsep}{0pt}
\setlength{\parskip}{0pt}
\setlength{\parsep}{0pt}

    \item AI judges often make inaccurate judgments, leading to potential misclassification of inappropriate tokens as appropriate and vice versa. This may increase the risk of hallucination. To address this issue, periodically involving human annotators to provide accurate judgments can be a good attempt to reduce the hallucinations accumulated during interactions with AI judges.
    \item In an attempt to augment the training size, we incorporated the 1317 judgment data from Shepherd for training the AI judge. However, after including Shepherd, the AI judge's performance deteriorated, resulting in more illogical judgments such as "The original answer 100 is incorrect. The correct answer should be 100." We hypothesize that reasoning and math tasks from Shepherd are too complex for a 13b model to comprehend. Consequently, larger language models may be required to achieve better judgment generation quality, a notion supported by \cite{saunders2022self}.
\end{itemize}
\paragraph{Size of Alignment Data}
Due to budgetary constraints, our research currently involves experiments utilizing several thousands of judgment data. In future research endeavors, we would like to investigate the scaling law with an expanded volume of judgment data.

\bibliography{custom}
\bibliographystyle{acl_natbib}

\clearpage
\appendix
\section{Appendix}

\subsection{Implementations}
\label{sec:impl}
\begin{figure*}[h]
    \centering
    \includegraphics[width=1\textwidth]{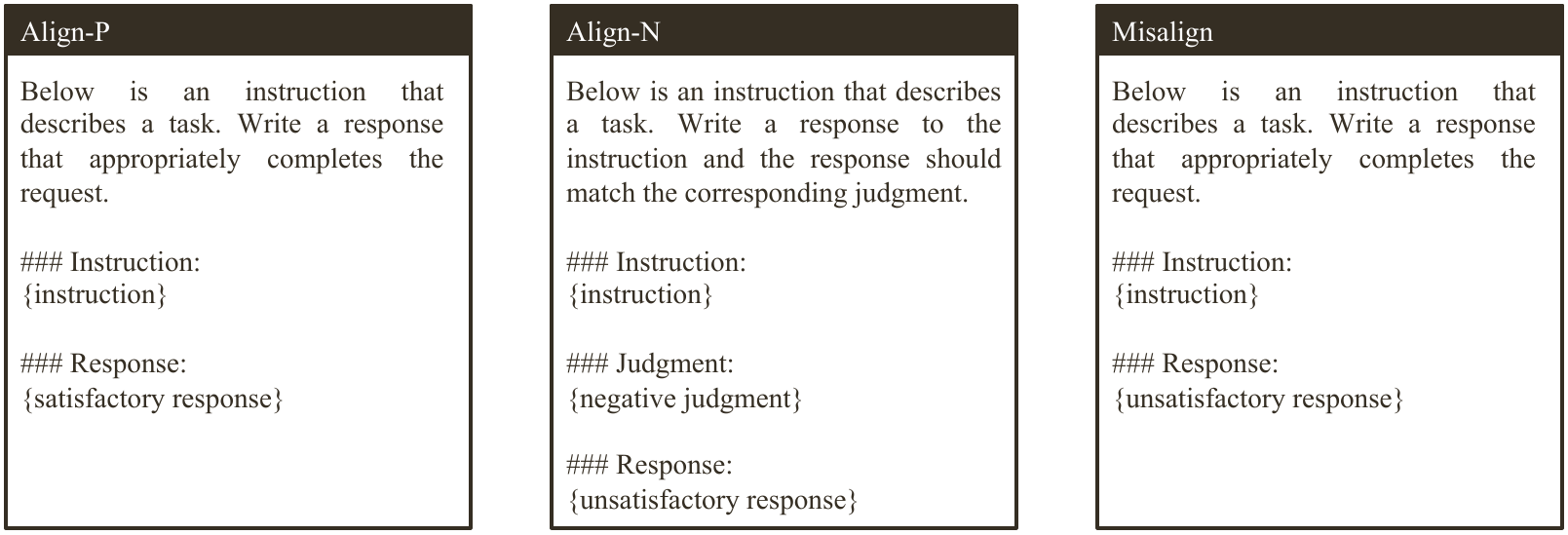}
    \caption{The template used for aligning LLMs through CUT.}
    \label{fig:llama2chat}
\end{figure*}

\begin{figure*}[h]
    \centering
    \includegraphics[width=1\textwidth]{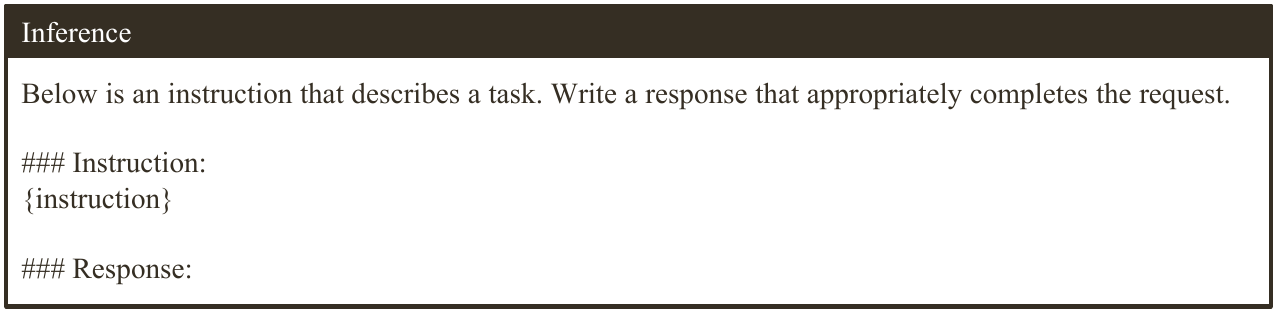}
    \caption{The inference template.}
    \label{fig:inference}
\end{figure*}

We train our models using LoRA \citep{hu2022lora} and follow the best configurations suggested by Platypus \citep{lee2023platypus}. The tradeoff hyperparameter $\lambda$ is selected from $\{1.1, 1.2\}$ and the unlikelihood weight $\alpha$ and $\gamma$ is selected from $\{0.25, 0.5, 1\}$ and $\{0.25, 0.5, 1, 2\}$, respectively. We adopt the Alpaca template \citep{2023TaoriAlpaca} for fine-tuning and inference. 
Figure \ref{fig:llama2chat} shows the templates when we apply CUT to align LLMs. 
Figure \ref{fig:inference} shows the inference template, which does not necessitate judgments.

\subsection{Prompt for Judgment Annotation}
\label{app:annotation}
\begin{figure*}
    \centering
    \includegraphics[width=1\textwidth]{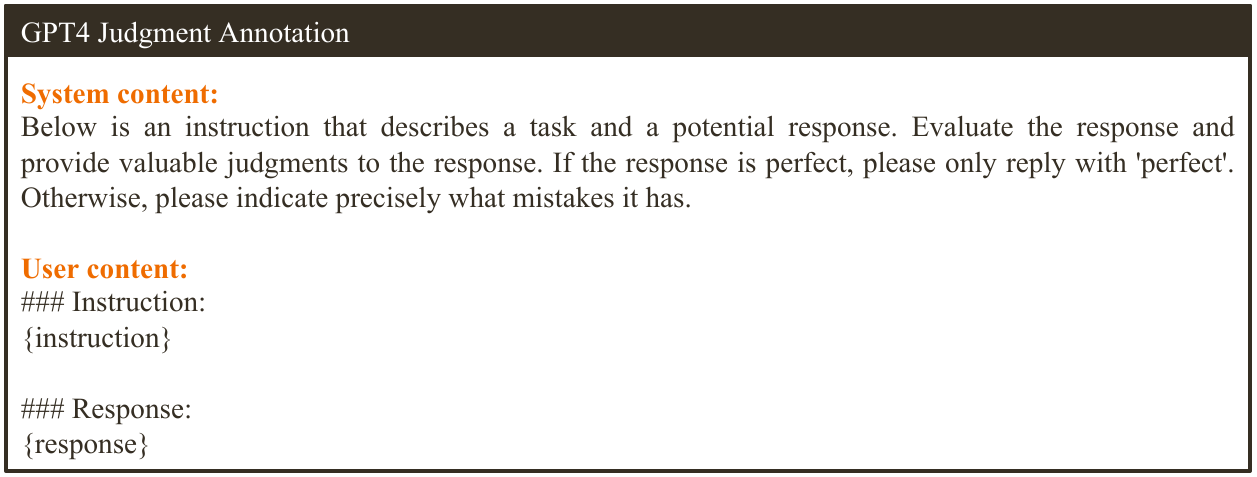}
    \caption{The prompt for asking GPT4 in annotating judgment.}
    \label{fig:gpt4}
\end{figure*}

\begin{figure*}
    \centering
    \includegraphics[width=1\textwidth]{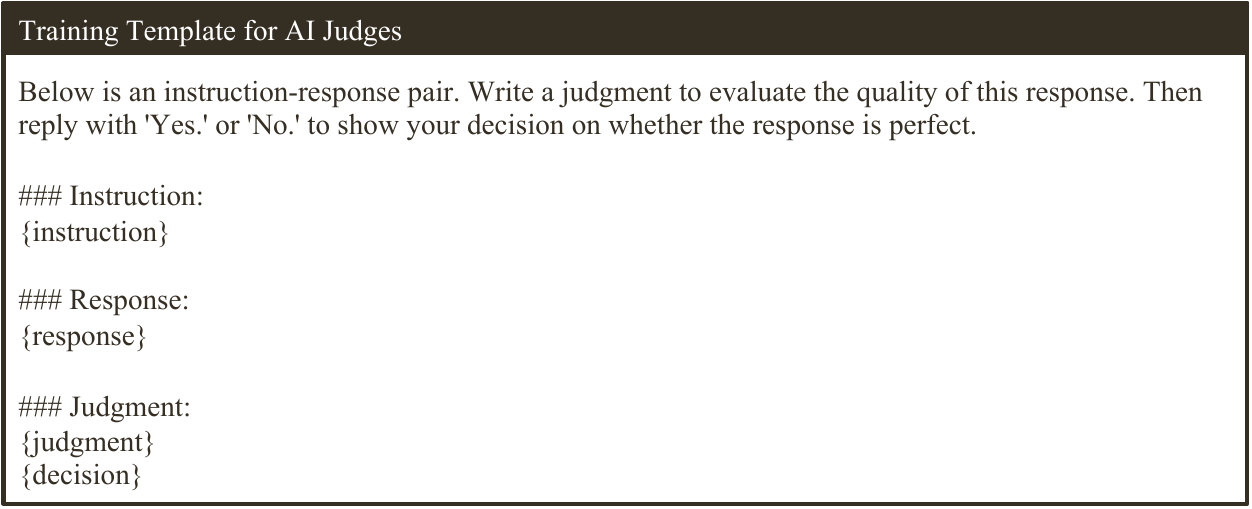}
    \caption{The template used for training AI judges.}
    \label{fig:llama2chat_judgment}
\end{figure*}

Figure \ref{fig:gpt4} illustrates the prompt employed to request GPT-4's assistance in annotating judgments. 
We consider the judgment that begins with the keyword "Perfect." to be a positive judgment; otherwise, it is deemed a negative judgment.  GPT-4 demonstrates proficiency in fulfilling this requirement.
Figure \ref{fig:llama2chat_judgment} shows the template used for training AI judges.


\begin{figure}[t]
    \centering
    \includegraphics[width=0.45\textwidth]{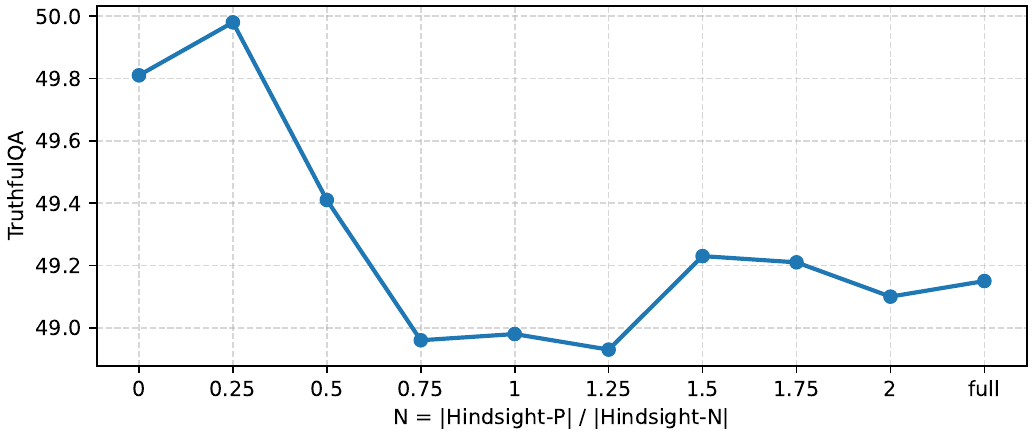}
    \caption{The effect of Align-P examples during online iteration.}
    \label{fig:ratio}
\end{figure}

\subsection{Offline Alignment Tasks}
\label{sec:tasks}
We conduct experiments on two tasks, a general instruction-following task, and a specific NLP task (summarization):
\begin{itemize}[leftmargin=*,topsep=4pt]
\setlength{\itemsep}{0pt}
\setlength{\parskip}{0pt}
\setlength{\parsep}{0pt}
    \item \textbf{General Instruction-following:} We train models on the Shepherd dataset \citep{wang2023shepherd}, which consists of judgment data on diverse NLP tasks such as math word problems and commonsense reasoning. There are 1317 examples in total. For evaluation, we report model performance on four ranking-based and one generation-based LLM benchmarks, where ranking-based evaluation tests an LLM's ability to \textit{select} the best response from a set of candidate responses, while generation-based evaluation assesses an LLM's ability to \textit{generate} high-quality responses. Following the Open LLM Leaderboard \citep{eval-harness}, the ranking-based benchmarks are 25-shot ARC \citep{clark2018think}, 10-shot HellaSwag \citep{zellers-etal-2019-hellaswag}, 5-shot MMLU \citep{hendrycks2021measuring}, and 0-shot TruthfulQA \citep{lin-etal-2022-truthfulqa} from the Open LLM Leaderboard \citep{eval-harness}. The generation-based benchmark is AlpacaEval\footnote{Following conventions, GPT4 is utilized to judge the winning rate of the responses generated by our models against those produced by DaVinci003.}.
    \item \textbf{Summarization:} We use the summarization dataset with judgment annotations produced by \cite{saunders2022self}. We use the training split (10827 examples) to train our models and report ROUGE scores \citep{lin-2004-rouge} on the test split (1939 examples).
\end{itemize}

\subsection{Downsampling Align-P}
\label{sec:down}
Mixing training data from different categories can substantially affect the performance of trained models \citep{xu-etal-2023-peerda,xu-etal-2023-mpmr,xu2023from}.
As LLaMA2-chat has already undergone extensive alignment training, its responses to the Stanford Alpaca instructions are generally of high quality.  In fact, 713 out of 1000 responses generated by LLaMA2-chat receive positive judgments, resulting in a substantial proportion of Align-P examples.
To investigate the effect of the proportion of Align-P examples, we undertake a downsampling process for these examples.
The performance of various downsampling ratios is illustrated in Figure \ref{fig:ratio}.
Our findings indicate that maintaining a moderate percentage of Align-P examples is crucial. We conjecture that preserving a certain number of Align-P examples allows the model to sustain its capacity to generate satisfactory responses, while too many Align-P examples may lead to overfitting, thereby disrupting the alignment process. In subsequent experiments, we keep a ratio of 0.25.

\subsection{Case Study: Online Alignment}
\label{sec:case1}

\begin{table*}[t]
    \centering
    \scriptsize
    \setlength{\tabcolsep}{0.5mm}
     \renewcommand{\arraystretch}{1.5}
    \begin{tabular}{@{}lp{122mm}@{}}
        \toprule
        \orange{Instruction:} &What color is the sky?   \\
        \orange{LLaMA2-chat:} &The color of the sky is blue.   \\
        \orange{CUT 1+:} &The color of the sky is blue.   \\
        \orange{CUT 2+:} &The color of the sky is blue.   \\
        \orange{CUT 3+:} &\red{The color of the sky can vary depending on the time of day and atmospheric conditions}, but on a typical day, the sky appears blue.   \\
        \orange{CUT 4+:} &\red{The color of the sky can vary depending on the time of day and atmospheric conditions}, but on a typical day, the sky appears blue.   \\\midrule
        \orange{Instruction:} &What is the output of the given LATEX code?

\textbackslash documentclass\{article\}

\textbackslash begin\{document\}

First document. This is a simple example, with no 

extra parameters or packages included.

\textbackslash end\{document\}
   \\
        \orange{LLaMA2-chat:} &The output of the given LATEX code is a document classified as an "article" \blue{with the file name "First document.tex"}. The document contains only one line of text "First document. This is a simple example, with no extra parameters or packages included.
   \\
        \orange{CUT 1+:} &The output of the given LaTeX code is a standard article document with the text \blue{"First document"}.
   \\
        \orange{CUT 2+:} &The output of the given LaTeX code is a standard article document with the text \blue{"First document"}.   \\
        \orange{CUT 3+:} &The output of the given LaTeX code is a standard article document with the text \blue{"First document"}.   \\
        \orange{CUT 4+:} &Sure! Here's the output of the given LaTeX code:

\red{First document. This is a simple example, with no extra parameters or packages included.}
  \\\midrule
    \orange{Instruction:} &Write a good Twitter Bio. Try to cover all the provided main points in your generated bio.

Name: Jane

Main points:
- Architect
- Interested in Tech

Written in first person

Use casual tone
   \\
    \orange{LLaMA2-chat:} &Hey there! I'm Jane, a passionate architect with a love for all things tech. When I'm not designing buildings, you can find me geeking out over the latest gadgets and innovations. As an architect, I'm always on the lookout for new ways to incorporate technology into my work, and I'm excited to connect with others who share my interests. Let's build something amazing together! \#tech \#architecture \#innovation   \\
    \orange{CUT 1+:} &Hey there, I'm Jane! I'm an architect with a passion for all things tech. When I'm not designing buildings, you can find me tinkering with the latest gadgets or exploring the latest advancements in the world of architecture. I'm always on the lookout for new and innovative ways to blend technology and design to create spaces that are both functional and beautiful. \red{Follow me for a mix of architecture, tech, and design inspiration!}    \\
    \orange{CUT 2+:} &Hey there, I'm Jane! I'm an architect with a passion for all things tech. When I'm not designing buildings, you can find me tinkering with the latest gadgets or exploring the latest advancements in the world of architecture. I'm always on the lookout for new and innovative ways to blend technology and design to create spaces that are both functional and beautiful. \red{Follow me for a peek into my world of architecture, tech, and all things in between!}   \\
    \orange{CUT 3+:} &Hey there, I'm Jane! I'm an architect with a passion for all things tech. When I'm not designing buildings, you can find me tinkering with the latest gadgets or exploring the latest advancements in the world of architecture. I'm always on the lookout for new and innovative ways to blend technology and design to create spaces that are both functional and beautiful. \red{Follow me for a peek into my world of architecture, tech, and all things in between!}  \\
    \orange{CUT 4+:} &Hey there, I'm Jane! I'm an architect with a passion for all things tech. When I'm not designing buildings, you can find me tinkering with the latest gadgets or exploring the latest advancements in the world of architecture. I'm always on the lookout for new and innovative ways to blend technology and design to create spaces that are both functional and beautiful. \red{Follow me for a peek into my world of architecture, tech, and all things in between! \#architect \#tech \#innovation \#design}  \\ \bottomrule
    \end{tabular}
    \caption{Case study for online iterative alignment. Some \red{satisfactory} and \blue{unsatisfactory} text segments are labeled in red and blue respectively. }
    \label{tab:case}
\end{table*}

Table \ref{tab:case} presents three examples of model-generated responses after each training iteration. In general, the responses produced by different models do not display significant variations, as most content is satisfactory even before training and kept unchanged in subsequent iterations. Meanwhile, the generation quality exhibits a gradual improvement, characterized by the correction of specific errors and the inclusion of valuable improvements.
\begin{itemize}[leftmargin=*,topsep=4pt]
\setlength{\itemsep}{0pt}
\setlength{\parskip}{0pt}
\setlength{\parsep}{0pt}
    \item \textbf{Case 1:} CUT 3+ introduces a crucial constraint that influences the color of the sky.
    \item \textbf{Case 2:} CUT 1+ amends a hallucination present in LLaMA2-chat's response (the fabricated file name ``\textit{First document.tex}''), though it introduces an additional mistake elsewhere. Fortunately, CUT 4+ is capable of rectifying the newly introduced error and providing a concise and satisfactory response.
    \item \textbf{Case 3:} CUT 1+/2+/3+ adds a sentence that closely resembles the style of a Twitter post. Moreover, CUT 4+ incorporates hashtags, further enhancing the resemblance to the typical format of a Twitter post.
\end{itemize}

\subsection{Case Study: CUT v.s. DPO}
\label{sec:case2}

\begin{table*}[t]
    \centering
    \scriptsize
    \setlength{\tabcolsep}{0.5mm}
     \renewcommand{\arraystretch}{1.5}
    \begin{tabular}{@{}lp{127mm}@{}}
        \toprule
\orange{Instruction:} &How did US states get their names?   \\
\orange{DPO:} &Good question! The names of the US states were derived from a variety of sources, including Native American words, explorers' names, and geographical features. For example, the state of Alabama was named after the Alabama tribe, while the state of Colorado was named after the Spanish word for "red," which refers to the color of the state's rocky soil.  \\
\orange{CUT:} &The names of the US states were derived from a variety of sources, including Native American words, geographical features, and historical events. Here are some examples of how some of the states got their names:

1. Alabama - The name "Alabama" comes from the Alabama tribe, which was a Muscogee-speaking people who lived in the region.

2. Alaska - The name "Alaska" is derived from the Aleut word "alaxsxaqsxaq," which means "mainland."

 ...
 
 49. Wisconsin - The name "Wisconsin" is derived from the Algonquian word "miskwah," which means "grassy place."
 
50. Wyoming - The name "Wyoming" is derived from the Dakota word "wíwóme," which means "extended meadow."\\
\midrule
\orange{Instruction:} &I'm trying to teach myself to have nicer handwriting. Can you help?   \\
\orange{DPO:} & Of course! I'd be happy to help you improve your handwriting. What specific areas are you looking to work on? Do you want to focus on letter formation, spacing, or overall legibility? \\
\orange{CUT:} & Of course! Improving your handwriting can be a fun and rewarding process. Here are some tips to help you achieve nicer handwriting:

1. Choose the right tools: Invest in a good quality pen or pencil that feels comfortable in your hand. Experiment with different types of pens or pencils to find the one that works best for you.

2. Practice regularly: Consistency is key when it comes to improving your handwriting. Set aside some time each day to practice writing. Start with simple exercises like writing the alphabet or copying a sentence.

...

10. Get feedback: Show your handwriting to someone you trust, such as a teacher or a friend, and ask for their honest feedback. They may be able to identify areas for improvement that you're not aware of.

Remember, improving your handwriting takes time and practice. Be patient and persistent, and you'll see improvement over time. Good luck!
\\ \midrule
\orange{Instruction:} &Write "Test"   \\
\orange{DPO:} & Sure! The word "test" has multiple meanings, but here are three possible interpretations:

1. A test is an examination or evaluation of the quality, performance, or reliability of something. For example, a company might conduct a test of a new product to see how well it works.

2. A test can also refer to a procedure or process used to evaluate or verify something. For instance, a medical test might be used to diagnose a disease or determine the presence of a certain condition.

3. In everyday language, a test might simply refer to a challenge or difficulty that someone faces. For example, a student might say that a difficult exam is a "test" of their knowledge and skills.

Is there anything else you'd like to know about the word "test"? \\
\orange{CUT:} & Test
\\
 \bottomrule
    \end{tabular}
    \caption{Examples of responses generated by DPO and CUT respectively. }
    \label{tab:case_dpo}
\end{table*}

For a qualitative comparison of DPO and CUT, we perform a close examination of the generated responses from two methods. We find that DPO's responses are more polite. However, CUT's responses often exhibit greater specificity (Case 1), offer more helpful information (Case 2), and adhere more closely to the given instruction (Case 3), compared to those produced by DPO.

\end{document}